\documentclass{amia}
\usepackage{graphicx}
\usepackage[labelfont=bf]{caption}
\usepackage[superscript,nomove]{cite}
\usepackage{color}
\usepackage{float}
\usepackage{amsmath}

\usepackage[hang]{footmisc}
\begin{document}

\title{Using Computer Vision to Automate Hand Detection and Tracking of Surgeon Movements in Videos of Open Surgery}


\author{Michael Zhang, BA$^{\dag 1}$\let\thefootnote\relax\footnotetext{$^\dag$Work done as visiting student researchers at Stanford University. \\
}, 
Xiaotian Cheng, BS$^{\dag 2}$, 
Daniel Copeland, BE$^{3}$, 
Arjun Desai, BS$^{4}$,\\
Melody Y. Guan, MA$^{5}$,
Gabriel A. Brat, MD MPH$^{\ddagger 3}$, 
Serena Yeung, PhD $^{\ddagger 6}$}

\institutes{
    $^{1}$Harvard University, Cambridge, MA; 
    $^{2}$Department of Automation, Tsinghua University, Beijing, China;
    $^{3}$Department of Surgery, Beth Isreal Deaconness Medical Center, Boston, MA;
    $^{4}$Department of Electrical Engineering, Stanford University, Stanford, CA
    $^{5}$Department of Computer Science, Stanford University, Stanford, CA
    $^{6}$Department of Biomedical Data Science, Stanford University, Stanford, CA\\
}

\maketitle


\noindent{\bf Abstract}

\textit{
Open, or non-laparoscopic surgery, represents the vast majority of all operating room procedures, but few tools exist to objectively evaluate these techniques at scale. Current efforts involve human expert-based visual assessment. We leverage advances in computer vision to introduce an automated approach to video analysis of surgical execution.  A state-of-the-art convolutional neural network architecture for object detection was used to detect operating hands in open surgery videos. Automated assessment was expanded by combining model predictions with a fast object tracker to enable surgeon-specific hand tracking. To train our model, we used publicly available videos of open surgery from YouTube and annotated these with spatial bounding boxes of operating hands. Our model's spatial detections of operating hands significantly outperforms the detections achieved using pre-existing hand-detection datasets, and allow for insights into intra-operative movement patterns and economy of motion.
}
\section{Introduction}
Improvements in surgical outcomes have been achieved by careful analysis of peri-operative hospitalization data to identify best practice and standardize care. Identification of evidence-based metrics has led to numerous pre- and post-surgery care pathways that significantly reduce re-admissions and morbidity. Despite such peri-operative standardization, significant differences remain in patient outcomes when stratified by surgeon\cite{Birkmeyer2013}. Accordingly, surgeon skill and conduct during an operation can significantly determine peri-operative outcome.

Despite these incentives, efforts to improve surgical quality are hampered by limited quantity and quality of data within the operative episode.  Surgical procedures are often merely recorded in retrospect by the practicing surgeon. These operative notes serve as the principle record of the surgery, but often poorly identify critical steps and overlook important aspects of individual procedures \cite{VanDeGraaf2019}.
Reliance on the subjective recall of surgeons also prevents effective cross-surgeon comparison and feedback on adherence to evidence-based practice.\cite{Makary2015}

Objective intra-operative data has the potential to generate new quality metrics and augment the capabilities of the surgeon\cite{Topol2019}. Various studies have demonstrated promising results using deep learning and computer vision for frame-level surgical tool detection in laparoscopic procedures and minimally invasive surgeries (MIS)\cite{raju2016m2cai, jin2018tool}. However, these surgeries represent less than 10 percent of all procedures. Open surgeries that require surgeons to manipulate tissues with their hands serve as the fundamental basis for many procedures today. Unfortunately, progress in automated assessment of open surgical skills$-$especially with vision-based approaches$-$is limited. There is an enormous gap in the ability to objectively evaluate intra-operative surgeon activity for these open approaches.\cite{Grenda2016}

\begin{figure}[t]
\centerline{
\includegraphics[width=17.5cm]{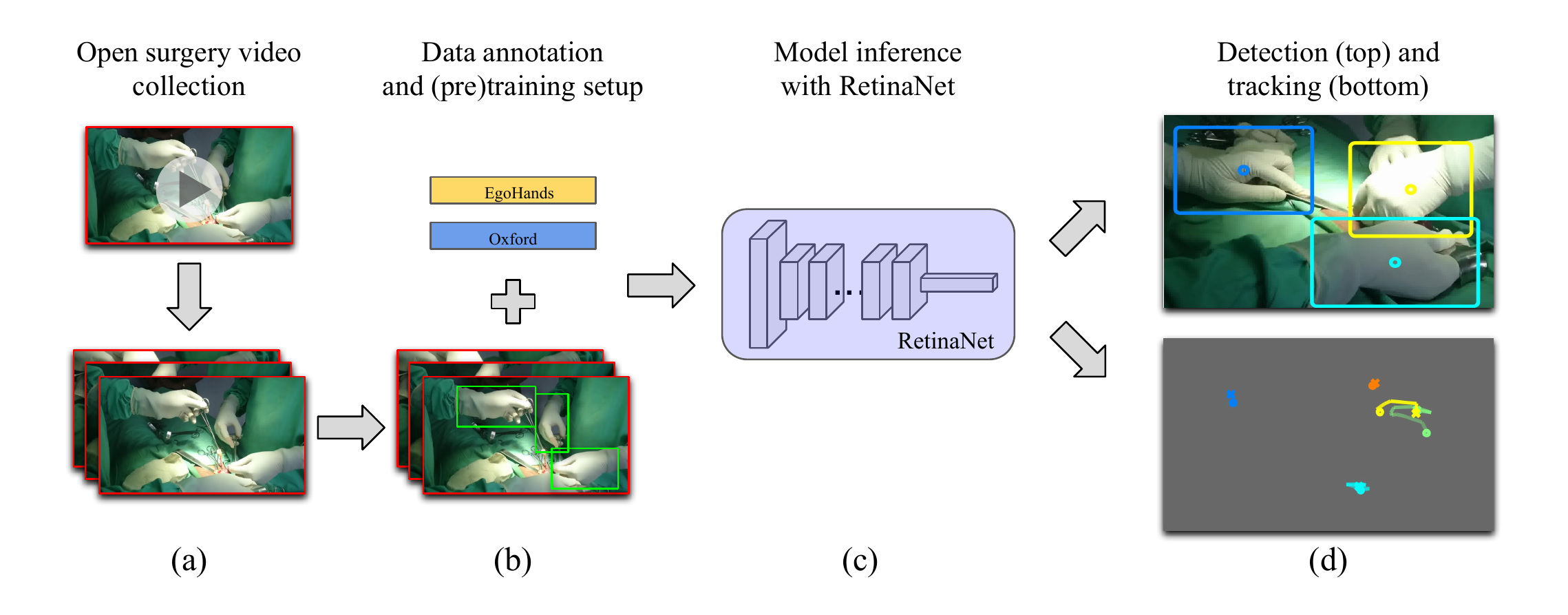}
}
\caption{In this work we propose a computer vision-based deep learning method to enable automated open surgery understanding. We collect a set of publicly available, open surgery videos from YouTube (a), and annotate video frames from these with spatial bounding boxes of operating hands to train a RetinaNet model for hand-specific object detection (b, c). 
Our model takes frames of unlabeled videos as input, and outputs inference for detection (d, top), which we follow with downstream hand-tracking (bottom) in surgeries.}
\label{aba:fig1}
\end{figure}

In this work, we take a significant step towards automated open surgery assessment with a computer vision-based deep learning method for detection of operating hands in surgery videos. While prior work involving supervised convolutional neural networks (CNNs) has shown promise for detecting tools, several conditions make the present task more difficult. Unlike tools, hands are visually deformable and may vary greatly in appearance. In addition, open surgery's enhanced generality$-$where surgeons may frequently manipulate several tools from multiple angles$-$along with variations in video quality, lighting, zoom levels, and camera angles present further challenges. Finally, while supervised deep learning models have demonstrated success in difficult object detection tasks taken from everyday settings\cite{DBLP:journals/corr/abs-1708-02002, DBLP:journals/corr/LinMBHPRDZ14}, they do so with significant quantities of diverse training data. Given the lack of obvious equivalents in open surgery, our problem additionally motivates the collection of similarly diverse training videos. 

We therefore combine a CNN framework for object detection with a diverse collection of open surgery videos obtained from YouTube, a publicly available  data source, which we annotate for the spatial bounding boxes of operating hands. Our object detection algorithm leverages \textit{RetinaNet}\cite{DBLP:journals/corr/abs-1708-02002}, a state-of-the-art CNN based on single-stage feature extraction and focal loss classification. Our approach achieves strong performance for detecting hands on a subset of the collected YouTube videos held-out for evaluation, substantially outperforming models with identical architectures trained on pre-existing hand-detection datasets. Finally, we show that combining our predicted detections with fast object-tracking algorithms enables temporally-consistent hand-tracking in surgery videos, allowing for further analysis regarding movement patterns and economy of motion to assess surgical performance.

\section{Related Work}
There has been a significant body of recent work targeting quantifiable and objective understanding of surgeries. The M2CAI 2016 Tool Presence Detection Challenge tasked participants to detect all surgical tools present in images taken from cholecystectomy procedures. Several deep learning architectures achieved state of the art performance \cite{raju2016m2cai}. Jin et al. extend this line of work by adding spatial tool localization on top of presence detection, and deploy their approach in videos to assess surgical performance \cite{jin2018tool}. Despite these advances, a majority of previous methods rely on laparoscopic or MIS procedures. In contrast, our work targets open surgeries, which remains a foundation of many surgical specialties but trails the former as an application domain in newer approaches for surgical understanding\cite{Shaharan17}. 

Assessment of open surgery is traditionally time-intensive and prone to human error, where industry standards revolve around checklist-based evaluations such as OSATS\cite{martin1997objective} and the watchful eye of a human expert. Accordingly, previous work on automated assessment includes a number of non-visual tracking approaches using either infrared light \cite{reiley2011review} or electromagnetic sensors placed on the surgeon's gloves\cite{dosis2005synchronized}. While there have been efforts to make these portable and practicable for motion tracking analysis \cite{Aggarwal2007}, the need for additional hardware introduces scalability issues and increased potential for hindering natural workflows. On the other hand, computer vision-based approaches hold the prospect of being comparatively easier to adopt and more readily available. One camera setup in an operating room can extend to track an arbitrary number of hands, unlike a wearable device that tracks a singular hand. Compared to sensors attached to instruments, vision-based systems are additionally relatively minimal and nonobtrusive \cite{reiley2011review}, two crucial properties shown to improve the efficiency of operating staff and tool usage\cite{riedl2002modern}. Finally, previous issues regarding camera-based tracking systems and line-of-sight obstructions\cite{Aggarwal2007} can be tackled with recent advances in computer vision. We demonstrate that our approach is robust to visual occlusion in many circumstances.

Outside of the surgical domain, there has been interest in hand-specific localization and tracking methods targeted to hands in the computer vision community. This has led to the release of datasets such as the EgoHands dataset \cite{Bambach_2015_ICCV}, which contains 4800 labeled video frames of hands from a first-person camera perspective, and the Oxford Hands dataset\cite{Mittal11}, which contained annotations for 13050 hand instances in third-person perspective images. While previous work in domain transfer might suggest that deep learning models trained on these diverse pre-existing datasets could then trivially be applied to track hands in the surgical setting, we empirically show that this is not the case. A separate line of work in computer vision has aimed to estimate entire body poses including hand keypoints.\cite{he2017mask, Guler2018DensePose} Although these do not explicitly target hand detection, in theory such methods could produce hand detection by calculating the tightest bounding box around all estimated hand keypoints. However, we found that these indirect approaches did not work well in practice on surgical videos. In our work, we therefore collect and annotate a new dataset of hands in videos of open surgery, and leverage a state-of-the-art object detection framework\cite{DBLP:journals/corr/abs-1708-02002} to train an effective model for detecting and tracking hands in surgical video.

\begin{figure}[t]
\centerline{
\includegraphics[width=18cm]{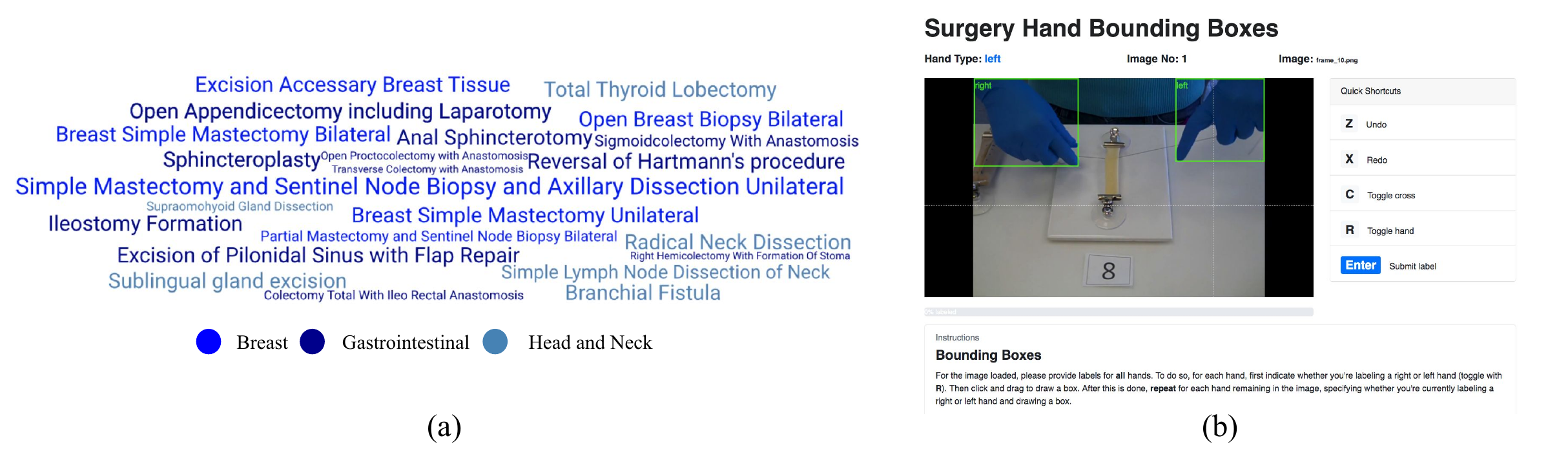}
}
\caption{(a) Cloud visualization of search terms represented in our collected videos, where font size is proportional to number of videos. (b) Interface for the labeling application we developed to collect annotations of operating hands.}
\label{data}
\end{figure}

\section{Approach}
In this section, we first describe the data collected for model development. We then present our model for hand detection in open surgery video frames, and describe how we link these detections across frames to track hand movement.

\subsection{Data}
In order to obtain a diverse set of data for our model development, we scraped videos of open surgery from YouTube. Specifically, we utilized a list of search terms compiled from all surgical procedures covered by a large insurance company (Southern Cross Health Society). Individual terms from the list were searched on YouTube (e.g. ``sublingual gland excision'') and queried results were added if they depicted open surgery, included surgeon hands, and were 144p or greater resolution. From all search terms in the breast, gastrointestinal, and head-and-neck surgery sections of the insurance company's list, 23 yielded at least one unique useful video (Fig. \ref{data}a). 
To train and evaluate our model on a good representation of surgeries, we then filtered our search to a dataset of 188 videos, consisting of 70 breast, 88 gastrointestinal and 30 head-and-neck videos.

We next selected a subset of all potential video frames to label with bounding box hand annotations for hand detection.
Videos were first converted to 15 fps. For videos over 20 minutes long, only the middle 20 minutes were processed. 
We then selected 10 frames at uniform intervals from each video, which allowed our image set to maintain diversity across different videos with respect to what kind of surgical procedure was being performed, the number of hand instances observed, and various video attributes such as quality, frame size, camera angle, perspective, and zoom.

To obtain annotations of the spatial boundaries of hands, we developed a web-based application allowing research assistants to draw bounding boxes for all hands in a given image (Figure \ref{data}b). 
The tool additionally asked annotators to indicate whether each hand was a right hand or a left hand; however for the purposes of this work, we did not use these annotations. Seven trained research assistants performed the labeling. Our final annotated dataset consists of 1880 labeled images, each containing zero or more bounding boxes associated with a hand.
Among the aggregated images, $940$ were allocated for training, $380$ for validation, and $560$ for testing. We split breast, gastrointestinal, and head-and-neck images equally across all sets, with no overlap in videos across sets. 


\begin{figure}[t]
\centerline{
\includegraphics[width=15.5cm]{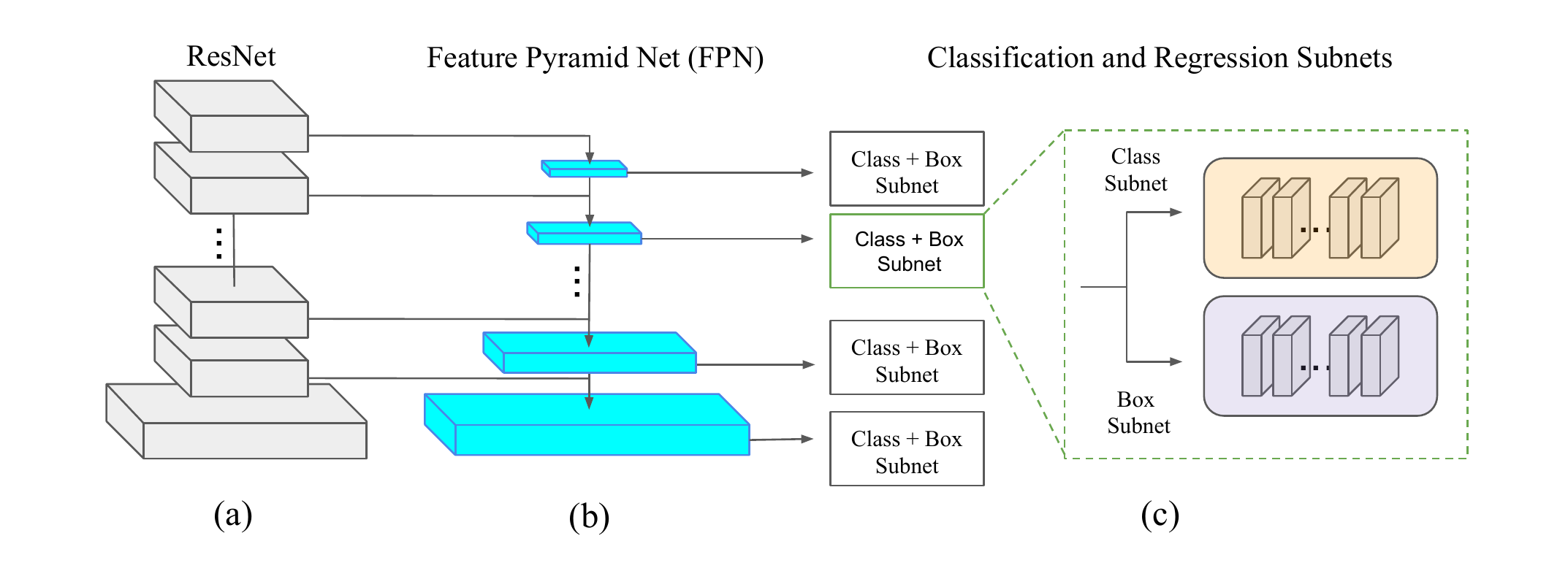}
}
\caption{RetinaNet architecture\cite{DBLP:journals/corr/abs-1708-02002}. Images are first fed into a standard feedforward ResNet model (a), which generates a multi-scale convolutional (conv) feature pyramid (b). At each level, RetinaNet carries connections to classification and bounding box subnets to classify and regress anchor boxes with respect to ground-truth object boxes (c). Subnets are composed of identical fully connected networks 
before a final 
conv layer with anchor filters and sigmoid activation.}
\label{retinanet}
\end{figure}

\subsection{RetinaNet Detection Model}
Our hand detection model is based on the RetinaNet\cite{DBLP:journals/corr/abs-1708-02002} neural network architecture, a CNN-based model for object detection that has achieved state-of-the-art performance on object detection tasks. We direct readers to Lin et al.\cite{DBLP:journals/corr/abs-1708-02002} for model details, but briefly describe RetinaNet's architecture as a single neural network composed of three parts: a \textit{backbone} responsible for converting image pixel representations to convolutional features, and two task-specific \textit{subnetworks} for classifying and regressing bounding boxes. The backbone incorporates a base Feature Pyramid Network (FPN)\cite{DBLP:journals/corr/LinDGHHB16}. This rests on top of a ResNet architecture which has been shown to extract powerful visual features\cite{DBLP:journals/corr/HeZRS15}. The FPN enables feature processing and object detection at multiple resolutions. Each level of the pyramid feeds into two fully-convolutional subnets, which aim to classify objects at every spatial position and output their spatial boundaries. 

To output predictions, RetinaNet divides input images according to a set of pre-defined reference boxes called ``anchors'', which correspond to sliding window positions of the backbone-generated feature map. Given an annotated image with a ground-truth bounding box, anchors are then assigned to the bounding box if the intersection-over-union (IoU) is greater than $0.5$, and to background if IoU $\in [0, 0.5)$. Image features are input to the subnetworks, which share the same architecture. We train the bounding box nets with standard $L2$ regression loss functions. 

In contrast to previous methods, RetinaNet uses a novel \textit{focal loss} for classification:
\begin{equation}
    \text{FL}(p_t) = -(1 - p_t)^\gamma \log(p_t),\;\; p_t = 
    \begin{cases}
        p &\text{if } y= 1 \\
        1-p &\text{otherwise}
    \end{cases}
\end{equation}
where $y \in \{-1, 1\}$ denotes a ground-truth class (e.g. if a hand exists or not) for an image representation, $p\in [0, 1]$ denotes the estimated probability for the class, and $(1 - p_t)^\gamma$ serves as a modulating factor with tunable focusing parameter $\gamma \geq 0$. Object detectors typically face a huge class imbalance prediction problem, where when trying to assign anchors to either target foreground classes or background, they may scan $10^4 - 10^5$ candidate locations per image only for a few to actually contain objects\cite{DBLP:journals/corr/abs-1708-02002}. Accordingly the $(1 - p_t)^\gamma$ tries to downweight easy examples, focusing training on hard negatives to maintain an optimal positive/negative ratio.

\subsection{Transfer Learning From Pre-existing Datasets}
While we anticipated that training on existing hand datasets alone would not transfer well to surgical hand detection due to large differences in domain appearance, we were interested in whether pre-training on them could help our model learn useful general hand detection representations, which could then be fine-tuned on our surgery video data. Accordingly, we experimented with various sequential pre-training  permutations. The datasets considered were:
\begin{enumerate}
    \item \textbf{EgoHands}\cite{bambach2015lending}, a collection of 4800 images from 48 videos containing pixel-level ground-truths for over 15 thousands hands. Videos depict various interactions between people, and each contribute 100 labeled images containing spatial boundaries of hands.
    \item \textbf{Oxford (Hands)}\cite{Mittal11}, a comprehensive compilation of $13050$ hand instances from various public image datasets. Of these, $4170$ are considered high quality hand instances with hands larger than 1500 square pixels.
\end{enumerate}

\subsection{Tracking Hands Across Frames}
Motivated by the importance of surgical assessment and correlation between metrics such as economy-of-motion with surgical skill and medical outcomes \cite{Birkmeyer2013}, we next apply the output of our hand detection model to frame-by-frame hand tracking in surgery videos. We use the \textit{Simple Online and Realtime Tracking} (SORT) algorithm, which enables multiple object tracking from un-identified bounding box inputs\cite{DBLP:journals/corr/BewleyGORU16}, and selected SORT for its simplicity and ability to perform quick inference for real-time tracking. SORT enables object-specific tracking through (1) object state propagation, (2) detection association with existing objects, and (3) lifespan management of tracked objects\cite{DBLP:journals/corr/BewleyGORU16}. During state propagation, SORT calculates metrics such as position, velocity, and box size given non-identified input bounding boxes and their spatial displacement across multiple frames, using a Kalman filter for motion prediction (1). Accordingly, for every updated prediction of each detection instance, SORT generates a predicted bounding box for the next frame, associating subsequent detection ground truths with the previous frame's boxes based on IoU and performing assignment given these metrics with the Hungarian algorithm (2). Finally to avoid unbounded growth of tracking identities, native SORT deletes identities every time objects move out of frame (3). However, due to the frequency at which hands belonging to the same surgeon may move in and out of sight in our dataset, we adapted SORT such that instead of deleting and creating a new identity upon an object's exit and re-entry into view, we merely update the pre-existing identity with the re-entry state calculated through (1), keeping track in a first-out, first-in manner. After model detections, we also augment SORT's tracking performance by interpolating predicted bounding boxes between frames during post-detection processing. Details are provided in Section 5.1 (Implementation Details).

\section{Experiments and Results}
We now evaluate our approach on detection and tracking of open surgery operating hands.
In Section $5.1$ we include further details on how we trained our model for detection and processed outputs for tracking. Section $5.2$ contains quantitative evaluations for our approach using the surgery hands dataset, and Section $5.3$ presents qualitative examples of hand tracking towards assessing surgical performance.

\subsection{Implementation Details}
For RetinaNet, we followed Lin et al. \cite{DBLP:journals/corr/abs-1708-02002}, where we used a ResNet-50-FPN backbone network, and an IoU threshold of $0.5$ between predicted bounding boxes and ground truth labels to denote a positive instance. All models were trained using the Adam optimizer with learning rate of $10^{-5}$ and batch size of $4$. Hyperparameters were compared on the validation set, and best parameters were selected to train on a combined training and validation set for the final model. For all datasets, we first trained our models for at least $50$ epochs or until convergence. For those pre-trained on multiple datasets, we trained sequentially, training until convergence completely with one dataset before moving on to another. At each stage, the best performing checkpoint was then selected for additional pre-training with a subsequent dataset or fine-tuning with our surgery dataset. For fine-tuning with the surgery dataset, we trained for $10$ epochs. Total training time for our largest dataset permutation took approximately five hours on an NVIDIA GeForce RTX 2080 Ti GPU. For post-detection processing, we found that a temporal window size of one frame (overall interpolation context of three frames), and a simple max-voting procedure to determine addition or subtraction of bounding boxes in the middle frame, led to the most stable SORT tracking output. After running test set frames through our hand detection model, we calculated interpolations with a stride-one sliding window for all frames from the same original video.

\subsection{Spatial Hand Detection in Open Surgery}
We compare hand detection performance using RetinaNet in Table 1. Training with our collected surgery hands dataset significantly outperforms training using existing hand datasets, and to the best of our knowledge our model is the first to demonstrate effective visual hand detection on real-world open surgery videos. All models were implemented with an identical RetinaNet architecture and tested against the same open surgery frames. For models trained using multiple datasets, we list the datasets in training order. We use mean average precision (mAP) to evaluate our model. A detection is considered correct if the intersection-over-union of a predicted bounding box with ground truth is at least $50\%$, i.e. an IOU threshold of 0.5.

Despite the vast quantity of annotated data in both the EgoHands and Oxford datasets, models only trained on these datasets perform substantially worse in comparison to those trained with our surgery data, suggesting a significant domain transfer problem related to the characteristics and representation of hands in a surgical environment. To improve performance, we explored pre-training with additional datasets, doing so sequentially with both the EgoHands and Oxford datasets. Because pre-training was done in succession, we also experimented with the order of training data. The model with our dataset remained ahead, but we found interestingly that on multiple occasions the biases present in the Oxford dataset seemed to be detrimental to model performance during fine-tuning.

\vspace{0.25cm}
\begin{table}[H]
\centering
\caption{Surgery detection performance across annotated hand datasets. Although hands are present in all datasets considered, training a RetinaNet model on existing datasets generalizes poorly to the surgery domain. We obtain a significant performance boost using our contributed data, demonstrating its value for detecting hands in surgery videos. }
\begin{tabular}{@{}lcccr@{}}
\hline
Training data (comma-separated by training order) & mAP ($\%$)\\ 
\hline
EgoHands & 11.8\\
Oxford & 3.4\\
EgoHands, Oxford & 5.9 \\
Oxford, EgoHands & 9.4\\
\textbf{Ours (surgery hands)} & \textbf{70.4}\\
\hline
\end{tabular}
\label{aba:tbl1}
\label{tab:mAP-test}
\end{table} 

\begin{table}[H]
\centering
\caption{Pretrained RetinaNet detection performance with fine-tuning on our dataset}
\begin{tabular}{@{}lcc@{}}
\hline
Pre-training dataset (comma-separated by training order) & mAP ($\%$)\\ \hline
None & 70.4\\
Oxford & 69.2\\
EgoHands & 74.8\\
Oxford, EgoHands & 73.7 \\
\textbf{EgoHands, Oxford} & \textbf{75.4}\\ 
\hline
\end{tabular}
\label{aba:tbl1}
\label{tab:mAP-test}
\end{table}

\vspace{-0.25cm}

We also investigated if pre-training on hand-detection datasets could lend useful detection priors for fine-tuning with our surgery hands dataset. Accordingly, we next evaluated performance using RetinaNet with our dataset and different pre-training permutations. 
We found that pre-training generally helped, and a permutation with all three datasets lead to
our best-observed performance of mAP$=75.4$ (Table 2). 

\begin{figure}[t]
\centerline{
\includegraphics[width=17.5cm]{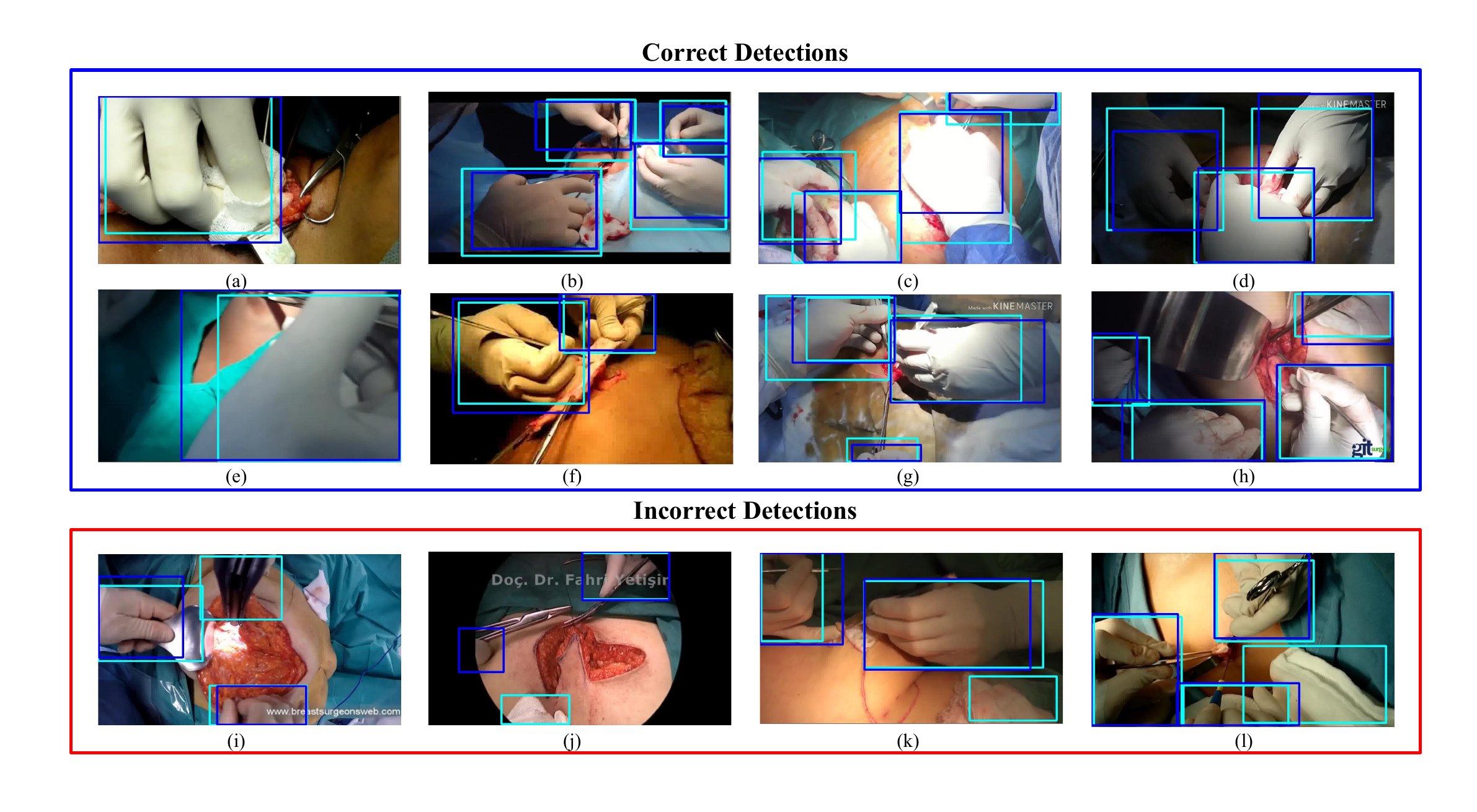}
}
\caption{Example hand detections on open surgery frames. Blue boxes represent ground truth, and cyan boxes denote predictions. We found the model to be effective across a variety of video qualities and camera perspectives, also successfully detecting multiple hands (b, c, d, f, g, h), and occluded instances or partial hands (c, e, g). Mistakes skewed towards false positives, with misclassification of objects appearing in plausibly similar contexts to hands.} 
\label{detections}
\end{figure}

Figure \ref{detections} contains example frames from our best-performing detection results, where our model successfully detects a variety of hand instances ranging in number, zoom, and image quality. Our model is effective at detecting hands in a variety of situations, including close-up shots (\ref{detections}a, \ref{detections}e), heavy occlusion by boundaries (\ref{detections}c, \ref{detections}g), and lighting variations (\ref{detections}d, \ref{detections}h). 
In aggregate our model suffered more from false positives, most prominently tending to misclassify objects that were both visually similar and appeared in positions or plausibly similar contexts to correct hands in the image (such as the cloth in Fig. \ref{detections}j, \ref{detections}k, \ref{detections}l). This suggests that our model learned more than hand-specific visual features to classify operating hands, and also provides one possible explanation to why visually-consistent hands in pre-existing datasets do not transfer well to the surgical domain.

\subsection{Assessing Surgical Performance with Automated Hand-tracking}
\begin{figure}[H]
\centerline{
\includegraphics[width=19cm]{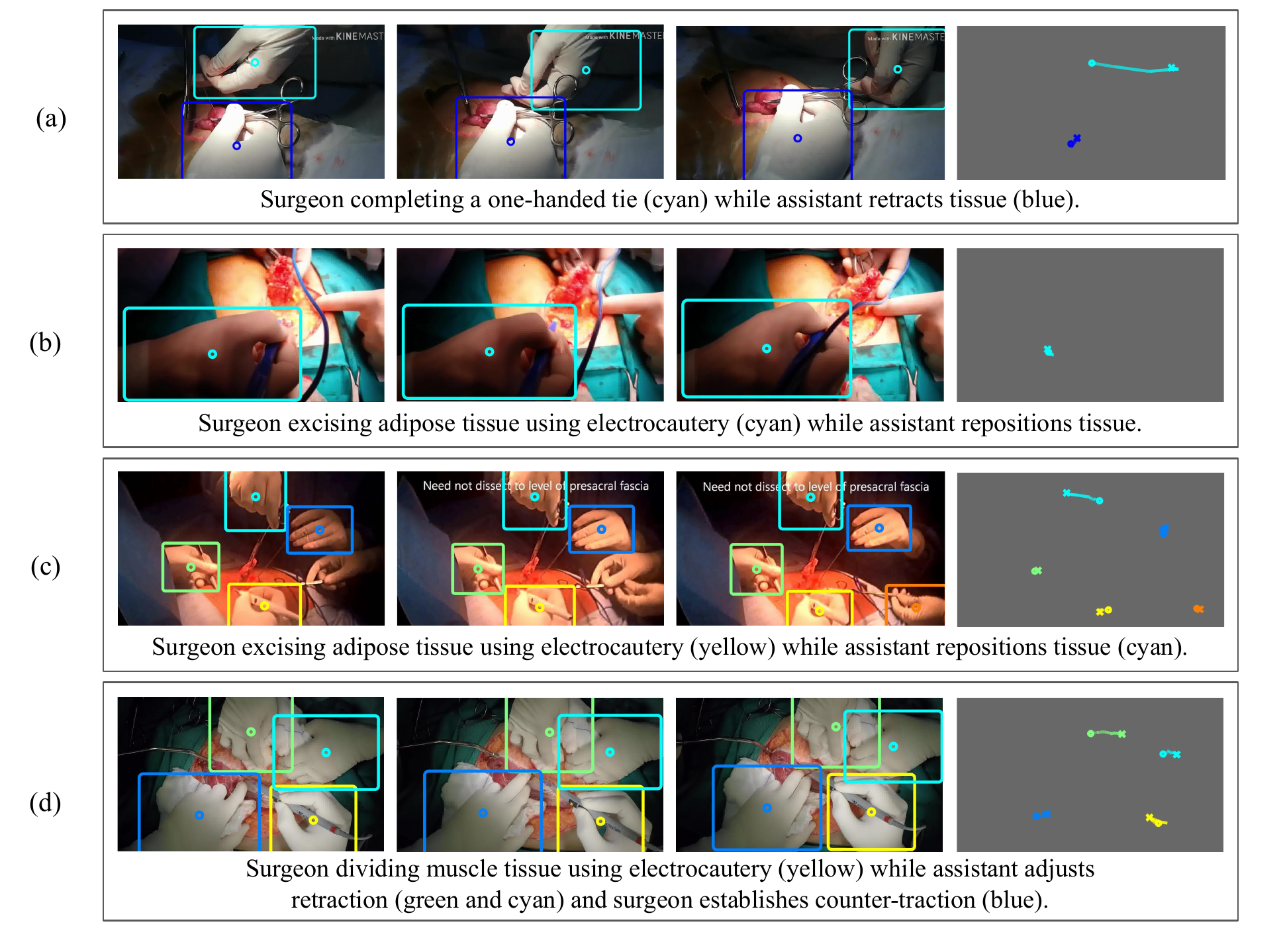}
}
\caption{Hand tracking during surgery videos. Figure is best viewed in color. Given post-processed predicted detection boxes as tracking algorithm input, we obtain temporally consistent and hand-specific frame-level assignments. Tracking boxes for videos are visualized on top of sequential frames sampled across time (first three columns). The epicenter of each box is used to generate trajectory maps depicting hand motion over time (last column). 
}
\label{tracking}
\end{figure}

We now assess the second part of our methods related to hand tracking. Given frame-by-frame model-predicted hand detections from our trained RetinaNet detector, we use SORT to generate hand-specific tracking predictions. Although we are unable to quantitatively assess tracking without ground-truth trajectory labels over sequences of video frames, we include qualitative examples of tracking outputs and their respective expert-driven interpretation in Figure \ref{tracking}, additionally creating multi-hand trajectory maps to aid in visualization (last column). These instances show that our tracking output is useful for assessing surgical characteristics such as motion pattern and economy of motion in various procedures. 

Our tracker was generally effective at identifying hands consistently through time, even in frames depicting multiple left and right hands (\ref{tracking}c, \ref{tracking}d). Additionally, we found that we could identify high economy of motion instances from the trajectory maps alone.
For example, actions requiring a steady hand such as excising tissue with an electrocautery rendered very little overall tracking movement (cyan in \ref{tracking}b, yellow in \ref{tracking}c, \ref{tracking}d). This was consistent across all four example videos. For techniques that involved larger lateral movements such as tying a knot with suture, trajectories were smooth and efficient (top hand, cyan in \ref{tracking}a). Finally, the mapped trajectories also highlight instances of highly controlled dexterity, such as in Figure \ref{tracking}d where the bottom left hand relies on minor finger adjustments as opposed to larger motions to apply counter-traction (blue), while the right hand uses electrocautery to divide tissue (yellow). This compares to the larger movements applied by assistants to ensure counter-traction (blue and green trajectories).
Without explicit hand identification labels, our detections and tracking algorithm were able to generate coherent hand-specific tracking. Overall, the trajectory maps highlighted insights that were consistent with an independent surgeon review of videos.

\section{Conclusion}
Towards automated assessment of open surgeries, we present a CNN-based computer vision model, trained on data of diverse, publicly accessible videos of open surgeries, which achieves strong performance on spatial hand detection in real-world surgeries. To the best of our knowledge, we are the first to produce an effective surgery hand detector from visual image-level data alone. Finally, we combine our detector with an off-the-shelf object tracking algorithm to enable hand-specific identification and tracking throughout videos, allowing further downstream assessment of crucial surgical properties such as movement patterns and economy of motion on singular hands. We hope to extend these capabilities in future work by further increasing the quantity and representation of surgery videos in our dataset, along with the addition of finer joint-specific labels and hand identification annotations.


\makeatletter
\renewcommand{\@biblabel}[1]{\hfill #1.}
\makeatother

\bibliographystyle{unsrt}
\bibliography{main.bib}

\begin{thebibliography}{10}

\bibitem{Birkmeyer2013}
John~D. Birkmeyer, Jonathan~F. Finks, Amanda O'Reilly, Mary Oerline, Arthur~M.
  Carlin, Andre~R. Nunn, Justin Dimick, Mousumi Banerjee, and Nancy~J.O.
  Birkmeyer.
\newblock {Surgical Skill and Complication Rates after Bariatric Surgery}.
\newblock {\em New England Journal of Medicine}, 369(15):1434--1442, 2013.

\bibitem{VanDeGraaf2019}
Floyd~W. {Van De Graaf}, Marilyne~M. Lange, Jolanda~I. Spakman, Wilhelmina~M.U.
  {Van Grevenstein}, Daan Lips, Eelco~J.R. {De Graaf}, Anand~G. Menon, and
  Johan~F. Lange.
\newblock {Comparison of Systematic Video Documentation with Narrative
  Operative Report in Colorectal Cancer Surgery}.
\newblock {\em JAMA Surgery}, 154(5):381--389, 2019.

\bibitem{Makary2015}
Martin~A. Makary, Tim Xu, and Timothy~M. Pawlik.
\newblock {Can video recording revolutionise medical quality?}
\newblock {\em BMJ (Online)}, 351(October):1--2, 2015.

\bibitem{Topol2019}
Eric~J Topol.
\newblock {High-performance medicine: the convergence of human and artificial
  intelligence}.
\newblock {\em Nature Medicine}.

\bibitem{raju2016m2cai}
Ashwin Raju, Sheng Wang, and Junzhou Huang.
\newblock M2cai surgical tool detection challenge report.
\newblock {\em University of Texas at Arlington, Tech. Rep.}, 2016.

\bibitem{jin2018tool}
Amy Jin, Serena Yeung, Jeffrey Jopling, Jonathan Krause, Dan Azagury, Arnold
  Milstein, and Li~Fei-Fei.
\newblock Tool detection and operative skill assessment in surgical videos
  using region-based convolutional neural networks.
\newblock In {\em 2018 IEEE Winter Conference on Applications of Computer
  Vision (WACV)}, pages 691--699. IEEE, 2018.

\bibitem{Grenda2016}
Tyler~R. Grenda, Jason~C. Pradarelli, and Justin~B. Dimick.
\newblock {Using surgical video to improve technique and skill}.
\newblock {\em Annals of Surgery}, 264(1):32--33, 2016.

\bibitem{DBLP:journals/corr/abs-1708-02002}
Tsung{-}Yi Lin, Priya Goyal, Ross~B. Girshick, Kaiming He, and Piotr
  Doll{\'{a}}r.
\newblock Focal loss for dense object detection.
\newblock {\em CoRR}, abs/1708.02002, 2017.

\bibitem{DBLP:journals/corr/LinMBHPRDZ14}
Tsung{-}Yi Lin, Michael Maire, Serge~J. Belongie, Lubomir~D. Bourdev, Ross~B.
  Girshick, James Hays, Pietro Perona, Deva Ramanan, Piotr Doll{\'{a}}r, and
  C.~Lawrence Zitnick.
\newblock Microsoft {COCO:} common objects in context.
\newblock {\em CoRR}, abs/1405.0312, 2014.

\bibitem{Shaharan17}
Shazrinizam Shaharan, Donncha~M Ryan, and Paul~C Neary.
\newblock Motion tracking system in surgical training.
\newblock In Carlos~M. Travieso-Gonzalez, editor, {\em Motion Tracking and
  Gesture Recognition}, chapter~1. IntechOpen, Rijeka, 2017.

\bibitem{martin1997objective}
JA~Martin, Glenn Regehr, Richard Reznick, Helen Macrae, John Murnaghan, Carol
  Hutchison, and M~Brown.
\newblock Objective structured assessment of technical skill (osats) for
  surgical residents.
\newblock {\em British journal of surgery}, 84(2):273--278, 1997.

\bibitem{reiley2011review}
Carol~E Reiley, Henry~C Lin, David~D Yuh, and Gregory~D Hager.
\newblock Review of methods for objective surgical skill evaluation.
\newblock {\em Surgical endoscopy}, 25(2):356--366, 2011.

\bibitem{dosis2005synchronized}
Aristotelis Dosis, Rajesh Aggarwal, Fernando Bello, Krishna Moorthy, Yaron
  Munz, Duncan Gillies, and Ara Darzi.
\newblock Synchronized video and motion analysis for the assessment of
  procedures in the operating theater.
\newblock {\em Archives of Surgery}, 140(3):293--299, 2005.

\bibitem{Aggarwal2007}
R.~Aggarwal, A.~Dosis, F.~Bello, and A.~Darzi.
\newblock Motion tracking systems for assessment of surgical skill.
\newblock {\em Surgical Endoscopy}, 21(2):339--339, Feb 2007.

\bibitem{riedl2002modern}
Stefan Riedl.
\newblock Modern operating room management in the workflow of surgery. spectrum
  of tasks and challenges of the future.
\newblock {\em Der Chirurg}, 73(2):105--110, 2002.

\bibitem{Bambach_2015_ICCV}
Sven Bambach, Stefan Lee, David~J. Crandall, and Chen Yu.
\newblock Lending a hand: Detecting hands and recognizing activities in complex
  egocentric interactions.
\newblock In {\em The IEEE International Conference on Computer Vision (ICCV)},
  December 2015.

\bibitem{Mittal11}
A.~Mittal, A.~Zisserman, and P.~H.~S. Torr.
\newblock Hand detection using multiple proposals.
\newblock In {\em British Machine Vision Conference}, 2011.

\bibitem{he2017mask}
Kaiming He, Georgia Gkioxari, Piotr Doll{\'a}r, and Ross Girshick.
\newblock Mask r-cnn.
\newblock In {\em Proceedings of the IEEE international conference on computer
  vision}, pages 2961--2969, 2017.

\bibitem{Guler2018DensePose}
Iasonas~Kokkinos Riza Alp~G\"uler, Natalia~Neverova.
\newblock Densepose: Dense human pose estimation in the wild.
\newblock 2018.

\bibitem{DBLP:journals/corr/LinDGHHB16}
Tsung{-}Yi Lin, Piotr Doll{\'{a}}r, Ross~B. Girshick, Kaiming He, Bharath
  Hariharan, and Serge~J. Belongie.
\newblock Feature pyramid networks for object detection.
\newblock {\em CoRR}, abs/1612.03144, 2016.

\bibitem{DBLP:journals/corr/HeZRS15}
Kaiming He, Xiangyu Zhang, Shaoqing Ren, and Jian Sun.
\newblock Deep residual learning for image recognition.
\newblock {\em CoRR}, abs/1512.03385, 2015.

\bibitem{bambach2015lending}
Sven Bambach, Stefan Lee, David~J Crandall, and Chen Yu.
\newblock Lending a hand: Detecting hands and recognizing activities in complex
  egocentric interactions.
\newblock In {\em Proceedings of the IEEE International Conference on Computer
  Vision}, pages 1949--1957, 2015.

\bibitem{DBLP:journals/corr/BewleyGORU16}
Alex Bewley, ZongYuan Ge, Lionel Ott, Fabio Ramos, and Ben Upcroft.
\newblock Simple online and realtime tracking.
\newblock {\em CoRR}, abs/1602.00763, 2016.

\end{thebibliography}



\end{document}